\documentclass[11pt]{article}
\usepackage[final]{acl}

\usepackage{times}
\usepackage{latexsym}
\usepackage[T1]{fontenc}
\usepackage[utf8]{inputenc}
\usepackage{microtype}
\usepackage{inconsolata}

\usepackage{graphicx}
\usepackage{booktabs}
\usepackage{multirow}
\usepackage{array}

\usepackage{amsmath}

\usepackage{algorithm}
\usepackage{algpseudocode}

\usepackage{xspace}
\usepackage{enumitem}
\usepackage{xcolor}
\usepackage{url}

\usepackage{xfp}
\newcommand{\maxgain}{7.0}
\newcommand{\gain}[1]{%
  \begingroup
  \edef\bg{\fpeval{round(15 + 70*(#1/\maxgain),0)}}%
  \setlength{\fboxsep}{0.6pt}%
  \colorbox{green!\bg}{\scriptsize\strut(+#1)}%
  \endgroup
}

\newcommand{\llamaThreeTwo}{Llama-3.2-3B-Instruct\xspace}
\newcommand{\llamaThreeOne}{Llama-3.1-8B-Instruct\xspace}

\title{POaaS: Minimal-Edit Prompt Optimization as a Service \\ to Lift Accuracy and Cut Hallucinations on On-Device sLLMs}

\author{
  \textbf{Jungwoo Shim}\quad
  \textbf{Dae Won Kim}\quad
  \textbf{Sun Wook Kim}\quad
  \textbf{Soo Young Kim} \\
  \textbf{Myungcheol Lee}\quad
  \textbf{Jae-geun Cha}\quad
  \textbf{Hyunhwa Choi}\thanks{Corresponding author.}
  \\ \\
  Electronics and Telecommunications Research Institute, Republic of Korea \\
  \texttt{\{right\_rain,won22,swkim99,sykim,mclee,jgcha,hyunwha\}@etri.re.kr}
}

\begin{document}
\setcounter{footnote}{1}
\maketitle
\setcounter{footnote}{0} 

\begin{abstract}
Small language models (sLLMs) are increasingly deployed on-device, where imperfect user prompts--typos, unclear intent, or missing context--can trigger factual errors and hallucinations. Existing automatic prompt optimization (APO) methods were designed for large cloud LLMs and rely on search that often produces long, structured instructions; when executed under an on-device constraint where the \emph{same} small model must act as optimizer and solver, these pipelines can waste context and even hurt accuracy. 
We propose \textbf{POaaS}, a minimal-edit prompt optimization layer that routes each query to lightweight specialists (Cleaner, Paraphraser, Fact-Adder) and merges their outputs under strict drift and length constraints, with a conservative skip policy for well-formed prompts. Under a strict fixed-model setting with \llamaThreeTwo{} and \llamaThreeOne{}, POaaS improves both task accuracy and factuality while representative APO baselines degrade them, and POaaS recovers up to \textbf{+7.4\%} under token deletion and mixup. Overall, per-query conservative optimization is a practical alternative to search-heavy APO for on-device sLLMs.
\end{abstract}

\section{Introduction}
\label{sec:intro}

Large language models (LLMs) have rapidly become the default interface for knowledge work, question answering, and decision support. Although early deployments relied on large, server-hosted models~\cite{leon2025gpt, sparkman2025claude}, recent advances are pushing \emph{small} language models (sLLMs) onto phones, browsers, and edge devices~\cite{gunter2024appleintelligencefoundationlanguage, grattafiori2024llama3herdmodels, roh2024towards}.

The first line of defense has been \emph{manual prompt engineering}~\cite{reynolds2021promptprogramming}, but it is labor-intensive and brittle, and small changes in domain, user intent, or input quality often require re-design.
To reduce this overhead, \emph{automatic prompt optimization} (APO) frameworks emerged~\cite{wan2024teach}. Systems such as OPRO~\cite{yang2023large}, PromptWizard~\cite{agarwal2025promptwizard}, and EvoPrompt~\cite{tong2025evoprompt} use iterative search and evolution to discover prompts that help large cloud LLMs. However, these assumptions mismatch on-device sLLMs. The search itself is expensive, optimized prompts can become long and structured, and increased prompt complexity can raise verbose eneration and hallucination risk in smaller models~\cite{li2025survey, ramnath2025systematic}. This motivates a different design point: \emph{per-query, conservative edits} that fix obvious input issues while avoiding unnecessary rewriting.

\paragraph{Our approach: POaaS.}
We introduce \textbf{Prompt Optimization as a Service (POaaS)}, a lightweight layer between the user prompt and the target model. POaaS uses three specialists--\textbf{Cleaner} (typos/grammar), \textbf{Paraphraser} (clarity/fluency), and \textbf{Fact-Adder} (concise contextual facts)--but applies them selectively via CPU-only routing. It can \emph{skip refinement} for well-formed prompts, and a drift-controlled merger rejects edits that deviate too far from the user’s intent. GPU time remains dominated by the specialist calls that are actually used.

\paragraph{Empirical findings.}
Under a strict fixed-model policy, we compare POaaS to APO baselines on reasoning tasks (BBH~\cite{suzgun2023challenging}, GSM8K~\cite{cobbe2021training}, CommonsenseQA~\cite{talmor2019commonsenseqa}) and factuality benchmark datasets (HaluEval~\cite{li2023halueval}, HalluLens~\cite{bang2025hallulens}, FActScore~\cite{min2023factscore}), under clean and degraded inputs. POaaS yields consistent gains on both \llamaThreeTwo{} and \llamaThreeOne{}, while APO baselines degrade performance and add large prompt overheads; under 5--15\% token deletion/mixup, POaaS recovers up to +7.4\% over No Optimization.

\paragraph{Contributions.}
    \begin{itemize}
        \item We introduce \textbf{POaaS}, a minimal-edit, microservice-style optimization layer for sLLMs that sits between user prompts and the target model, and show that it consistently improves accuracy where state-of-the-art APO methods reduces instead.
        \item We show that POaaS is particularly effective under realistic input degradations that mimic human error (typos, deletions, and missing information), recovering performance on both reasoning and factuality benchmarks where APO baselines further amplify degradation.
        \item We provide a capacity-aware design to deliver these gains with zero offline optimization cost and modest per-query latency, making POaaS suitable for on-device and edge deployments.
    \end{itemize}


\section{Related Work}
\label{sec:related}

\subsection{Automatic Prompt Optimization (APO)}
\label{sec:apo}

Automatic prompt optimization methods treat prompts as objects of search, typically optimizing performance through iterative rewriting, scoring, and selection~\citep{ramnath2025systematic}. A recent survey characterizes APO along dimensions such as candidate generation, evaluation strategy, and search depth, spanning gradient-free methods, meta-prompt design, and program-synthesis-style pipelines~\cite{chang2024efficient}.

Representative systems include OPRO, which conditions on previously generated prompts and their scores in a meta-prompt to propose new instructions; PromptWizard, which uses a self-evolving loop of LLM-driven mutation, critique, and in-context example optimization; and EvoPrompt, which connects LLMs with evolutionary algorithms implementing mutation and crossover operators. Other variants explore self-referential prompt evolution (PromptBreeder~\cite{fernando2023promptbreeder}), gradient-like textual feedback with bandit-style candidate selection and Monte Carlo search (ProTeGi~\cite{pryzant2023automatic}), Monte Carlo Tree Search–based strategic planning over prompt states (PromptAgent~\cite{wang2023promptagent}), and pipeline-level prompt and weight optimization for multi-stage LM programs (DSPy~\cite{khattab2024dspy}).

These methods have shown strong gains on cloud-scale LLMs, but their assumptions differ from our target setting. They typically evaluate many candidates, require substantial compute, and often yield long, highly structured prompts that consume scarce tokens on small models. Moreover, most APO work focuses on task accuracy rather than hallucination robustness, and the resulting prompt complexity can increase over-generation in sLLMs~\cite{cui2025automatic}. 

\subsection{Hallucination Mitigation Prompting}
\label{sec:prompt-hallucinations}

Hallucinations---fluent, but unsupported statements---pose critical reliability risks, especially for smaller models with limited contextual and reasoning capacity. Several benchmarks provide complementary views of this phenomenon: FActScore decomposes generations into atomic claims and measures evidential support; HaluEval offers a diverse hallucination test suite with LLM-as-judge evaluation; and HalluLens organizes intrinsic vs.\ extrinsic hallucinations and provides fine-grained error profiles. 

Most hallucination mitigation techniques operate \emph{after} generation, using retrieval-augmented generation (RAG)~\cite{lewis2020retrieval}, verifier models, or domain guardrails. While effective for large server-hosted LLMs, these methods increase system complexity and latency and may be impractical for resource-constrained sLLM deployments. By contrast, much less work focuses on \emph{pre}-generation mitigation at the prompt level, such as cleaning noisy inputs, clarifying underspecified instructions, or injecting lightweight context before decoding. Such approaches are attractive for on-device or latency-sensitive settings, because they can wrap a fixed sLLM with a small controller and compose with RAG or verification when available. 

Our work follows this direction: we view hallucination risk as partly induced by imperfect user prompts and introduce a modular prompt-optimization layer that edits inputs within a fixed token budget, positioning prompt-level optimization as a complementary building block alongside post-hoc retrieval and verification.

\begin{figure*}
    \centering
    \includegraphics[width=1\linewidth]{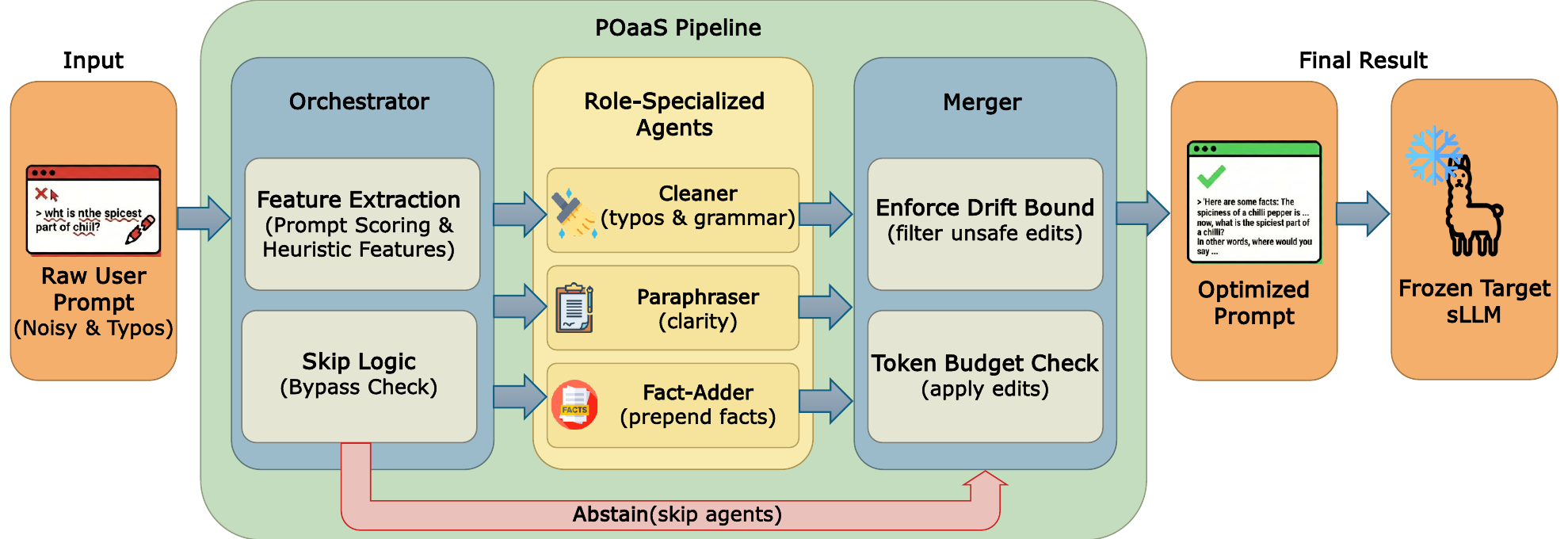}
    \caption{POaaS pipeline overview. First, the orchestrator analyzes input prompts using lightweight heuristics, routes to appropriate specialists (Cleaner, Paraphraser, Fact-Adder), and then the Merger applies drift-controlled merging to produce optimized prompts.}
    \label{fig:pipeline}
\end{figure*}
\begin{algorithm}[t]
\caption{POaaS Minimal-Edit Optimization Pipeline}
\label{alg:POaaS}
\footnotesize
\begin{algorithmic}[1]
  \Require Prompt $x$; routing thresholds
           $\tau_{\text{typo}}, \tau_{\text{comp}}, \tau_{\text{flu}}, \tau_{\text{skip}}$
  \Require Drift/length caps
           $\delta_{\text{clean}}, \delta_{\text{para}}, \delta_{\max}$; length cap $\rho_{\max}$
  \Require Drift params
          $\theta_{\text{drift}} =
            (\delta_{\text{clean}}, \delta_{\text{para}},
             \delta_{\max}, \rho_{\max})$
  \Ensure Optimized prompt $\tilde{x}$

  \State $\phi \gets \textsc{AnalyzePrompt}\big(
        x,\tau_{\text{typo}},\tau_{\text{comp}},
        \tau_{\text{flu}},\tau_{\text{skip}}\big)$

  \If{$\textsc{ShouldSkip}(\phi)$}
      \State \Return $x$
  \EndIf

  \State $A \gets \textsc{SelectAgents}(\phi)$
  \Comment{Selected agents}

  \State $\mathcal{C} \gets \emptyset$
  \Comment{Candidates}

  \For{each $a \in A$}
      \State $x' \gets a(x)$
      \State $\textit{ok} \gets \textsc{WithinDrift}(x, x'; \theta_{\text{drift}})$
      \If{$\textit{ok}$}
          \State $\mathcal{C} \gets \mathcal{C} \cup \{(a, x')\}$
      \EndIf
  \EndFor

  \If{$\mathcal{C} = \emptyset$}
      \State \Return $x$
  \EndIf

  \State $\tilde{x} \gets \textsc{Merge}(x, \mathcal{C})$
  \Comment{Compose edits + context}

  \State \Return $\tilde{x}$
\end{algorithmic}
\end{algorithm}

\section{Method}
\label{sec:method}

POaaS places a lightweight, minimal-edit optimization layer in front of a frozen target sLLM $f_\theta$ (Llama-3.2-3B-Instruct and Llama-3.1-8B-Instruct; decoding fixed across all runs: temperature $0.2$, top-$p$ $0.9$, fixed seed; max generation budget). Given a prompt $x$, POaaS (i) scores prompt quality with CPU-only heuristics and may skip optimization, (ii) selectively calls role-specialized agents (Cleaner, Paraphraser, Fact-Adder), and (iii) merges their outputs under strict drift/length and safety guards to produce $\tilde{x}$, which is sent to $f_\theta$. We illustrate POaaS's pipeline overview in Figure~\ref{fig:pipeline}.

\subsection{Task Setup}

Given a prompt $x$ and frozen $f_\theta$, POaaS seeks an optimized prompt $\tilde{x}$ that improves downstream utility while avoiding intent drift. We enforce:

\paragraph{Drift bound.}
We use a CPU-only lexical similarity ensemble and define
\begin{equation}
  D(x,x') = 1-\text{sim}(x,x') \in [0,1],
\end{equation}
accepting edits only if $D(x,x') \le \delta$ (details and ablations in Appendix~\ref{sec:app-drift}).

\paragraph{Length cap.}
We cap expansion by the character-length ratio
\begin{equation}
  \rho(x,\tilde{x}) = \frac{\text{len}(\tilde{x})}{\text{len}(x)} \le \rho_{\max},
\end{equation}
and also cap specialist outputs (Fact-Adder: $\le$120 tokens, up to 3 facts). Full budgets appear in Appendix~\ref{sec:app-budgets}.

\subsection{Prompt Analysis and Routing}
\label{sec:routing}

The orchestrator computes lightweight CPU scores (typo, completeness, fluency, clarity; all normalized to $[0,1]$) and applies threshold routing: Cleaner for high typo, Fact-Adder for low completeness, and Paraphraser for low fluency (Appendix~\ref{sec:app-scores}). Critically, POaaS uses a conservative \emph{skip} gate: if the prompt is already high-quality, POaaS returns $x$ unchanged to avoid harmful over-editing.

\paragraph{Skip logic.}
We compute an overall quality score
\begin{multline}
q(x)=1-\max\Big(\text{typo}(x),\,[\tau_{\text{comp}}-\text{comp}(x)]_+,\\
[\tau_{\text{flu}}-\text{flu}(x)]_+,\,[0.70-\text{clar}(x)]_+\Big),
\end{multline}
and skip refinement when $q(x) > 1-\tau_{\text{skip}}$ and typos are low, with defaults $\tau_{\text{skip}}=0.25$ (so $q(x)>0.75$) and $\text{typo}(x)<0.20$. This design is intentionally conservative: POaaS prefers false negatives (missing a small potential improvement) over false positives (unnecessary edits) to preserve intent on already well-formed prompts.

\subsection{Role-Specialized Agents}
\label{sec:agents}

All agents run as vLLM-served~\cite{kwon2023efficient} LoRA adapters~\cite{hu2021lora} on frozen Llama backbones. Each specialist is trained to produce \emph{one} transformation and then a secondary guard model verifies faithfulness:

\begin{itemize}[leftmargin=1.5em]
    \item \textbf{Cleaner} (fine-tuned with JFLEG~\cite{napoles2017jfleg}): fixes typos/grammar with a minimal-change instruction. A guard rejects candidates that introduce new information, reverting to the original prompt if unsafe. Examples and guard details appear in Appendix~\ref{sec:app-guards}.
    \item \textbf{Paraphraser} (fine-tuned with PAWS~\cite{zhang2019paws} and QQP~\cite{sharma2019qqp}): rewrites for fluency while preserving meaning. A fidelity guard rewrites or vetoes unfaithful paraphrases. Examples and thresholds appear in Appendix~\ref{sec:app-guards}.
    \item \textbf{Fact-Adder} (fine-tuned with Wikipedia and Wikidata-derived corpora~\cite{vrandevcic2014wikidata}): generates up to \textbf{three} concise factual bullets (total $\le$ \textbf{120 tokens} under the target tokenizer) that are directly related to the user’s input entities/task, providing lightweight contextual support for the downstream sLLM. If no high-confidence facts can be produced, it outputs \texttt{NONE} and is skipped during merging. A grounding/answer-leakage guard rejects reasoning-like content and removes unsupported statements before any fact is prepended.
\end{itemize}

\subsection{Agent Selection}
\label{sec:agent-selection}

Given the four scores from \S\ref{sec:routing}, POaaS uses thresholded routing:
Cleaner if $\text{typo}(x)>\tau_{\text{typo}}$; Fact-Adder if $\text{comp}(x)<\tau_{\text{comp}}$ (underspecified prompts); and Paraphraser if $\text{flu}(x)<\tau_{\text{flu}}$. Multiple agents may be invoked in parallel; their outputs are accepted only if they satisfy drift/length and safety guards, otherwise they are discarded and POaaS falls back to the original prompt.

\subsection{Drift-Controlled Merging}
\label{sec:merger}

POaaS merges specialist edits only when they are proven to be low-risk under cheap, deterministic checks. Concretely, each candidate $x'$ is sanitized (meta-commentary removed), scored for drift $D(x,x')$, checked for preservation of key content (entities/numbers/quotes), and rejected if it violates drift/length caps. Accepted edits are then composed into a single $\tilde{x}$ with a fixed precedence: apply Cleaner edits first, then Paraphraser, and finally prepend up to three Fact-Adder facts ($\le$120 tokens total) ahead of the user request. For few-shot prompts, POaaS isolates and edits only the final question span, reattaching the untouched prefix.

\paragraph{Safety and guards.}
POaaS uses layered fail-safes: per-agent guard models to reject unfaithful edits (adding facts, dropping constraints, or answering the question), orchestrator-side sanitization that strips meta-commentary, and strict structural preservation (e.g., few-shot exemplars are kept intact by editing only the final query span). If all candidates fail checks, POaaS returns the original prompt.

\paragraph{Implementation.}
POaaS is implemented as a FastAPI~\cite{lubanovic2023fastapi} orchestrator (CPU) that routes to vLLM+LoRA specialist services (GPU) and logs per-stage latency/tokens with deterministic configs for reproducibility are listed in Appendix~\ref{sec:app-instrumentation}.

\section{Experiments}
\label{sec:experiments}

We evaluate \textbf{POaaS} against state-of-the-art APO baselines on both task accuracy and factuality metrics, under clean and degraded input conditions. All experiments use fixed target models (Llama-3.2-3B-Instruct and Llama-3.1-8B-Instruct) with frozen weights and identical decoding parameters (temperature=0.2, top-$p$=0.9, max tokens= 512), ensuring that any differences are attributable solely to prompt optimization rather than model or decoding changes.

\subsection{Experimental Setup}
\label{subsec:setup}

\begin{table*}[t]
\centering
\caption{Task accuracy and factuality benchmark results (\%) for POaaS and APO baselines. \textcolor{green!70!black}{Green}/\textcolor{red!70!black}{red} show change vs No Optimization. Best in \textbf{bold}.}
\label{tab:main-results}
\resizebox{\textwidth}{!}{%
\begin{tabular}{llccc|c|ccc|c}
\toprule
& & \multicolumn{4}{c|}{\textbf{Task Accuracy}} & \multicolumn{4}{c}{\textbf{Factuality}} \\
\cmidrule(lr){3-6} \cmidrule(lr){7-10}
\textbf{Model} & \textbf{Method} & \textbf{BBH} & \textbf{GSM8K} & \textbf{CSQA} & \textbf{Avg.} & \textbf{HaluEval} & \textbf{HalluLens} & \textbf{FActScore} & \textbf{Avg.} \\
\midrule
\multirow{5}{*}{\textbf{Llama-3.2-3B}} 
& No Optimization & {42.2} & {77.2} & {71.6} & {63.7} & {68.2} & {48.8} & {16.6} & {44.5} \\
& EvoPrompt & 35.8 \textcolor{red!70!black}{\scriptsize(-6.4)} & 75.4 \textcolor{red!70!black}{\scriptsize(-1.8)} & 68.2 \textcolor{red!70!black}{\scriptsize(-3.4)} & 59.8 & 67.6 \textcolor{red!70!black}{\scriptsize(-0.6)} & 46.8 \textcolor{red!70!black}{\scriptsize(-2.0)} & 16.4 \textcolor{red!70!black}{\scriptsize(-0.2)} & 43.6 \\
& OPRO & 32.4 \textcolor{red!70!black}{\scriptsize(-9.8)} & 74.2 \textcolor{red!70!black}{\scriptsize(-3.0)} & 65.8 \textcolor{red!70!black}{\scriptsize(-5.8)} & 57.5 & 66.8 \textcolor{red!70!black}{\scriptsize(-1.4)} & 45.6 \textcolor{red!70!black}{\scriptsize(-3.2)} & 15.8 \textcolor{red!70!black}{\scriptsize(-0.8)} & 42.7 \\
& PromptWizard & 10.4 \textcolor{red!70!black}{\scriptsize(-31.8)} & 72.4 \textcolor{red!70!black}{\scriptsize(-4.8)} & 63.6 \textcolor{red!70!black}{\scriptsize(-8.0)} & 48.8 & 65.2 \textcolor{red!70!black}{\scriptsize(-3.0)} & 43.8 \textcolor{red!70!black}{\scriptsize(-5.0)} & 14.8 \textcolor{red!70!black}{\scriptsize(-1.8)} & 41.3 \\
& \textbf{POaaS (Ours)} & \textbf{46.0} \textcolor{green!70!black}{\scriptsize(+3.8)} & \textbf{79.0} \textcolor{green!70!black}{\scriptsize(+1.8)} & \textbf{73.0} \textcolor{green!70!black}{\scriptsize(+1.4)} & \textbf{66.0} & \textbf{70.2} \textcolor{green!70!black}{\scriptsize(+2.0)} & \textbf{52.0} \textcolor{green!70!black}{\scriptsize(+3.2)} & \textbf{22.0} \textcolor{green!70!black}{\scriptsize(+5.4)} & \textbf{48.1} \\
\midrule
\multirow{5}{*}{\textbf{Llama-3.1-8B}} 
& No Optimization & {51.8} & {82.4} & {76.2} & {70.1} & {74.6} & {56.2} & {24.8} & {51.9} \\
& EvoPrompt & 44.2 \textcolor{red!70!black}{\scriptsize(-7.6)} & 80.6 \textcolor{red!70!black}{\scriptsize(-1.8)} & 73.4 \textcolor{red!70!black}{\scriptsize(-2.8)} & 66.1 & 74.0 \textcolor{red!70!black}{\scriptsize(-0.6)} & 54.2 \textcolor{red!70!black}{\scriptsize(-2.0)} & 24.4 \textcolor{red!70!black}{\scriptsize(-0.4)} & 50.9 \\
& OPRO & 40.6 \textcolor{red!70!black}{\scriptsize(-11.2)} & 79.2 \textcolor{red!70!black}{\scriptsize(-3.2)} & 70.8 \textcolor{red!70!black}{\scriptsize(-5.4)} & 63.5 & 73.0 \textcolor{red!70!black}{\scriptsize(-1.6)} & 52.8 \textcolor{red!70!black}{\scriptsize(-3.4)} & 23.6 \textcolor{red!70!black}{\scriptsize(-1.2)} & 49.8 \\
& PromptWizard & 18.6 \textcolor{red!70!black}{\scriptsize(-33.2)} & 77.8 \textcolor{red!70!black}{\scriptsize(-4.6)} & 68.2 \textcolor{red!70!black}{\scriptsize(-8.0)} & 54.9 & 71.4 \textcolor{red!70!black}{\scriptsize(-3.2)} & 50.2 \textcolor{red!70!black}{\scriptsize(-6.0)} & 22.6 \textcolor{red!70!black}{\scriptsize(-2.2)} & 48.1 \\
& \textbf{POaaS (Ours)} & \textbf{54.0} \textcolor{green!70!black}{\scriptsize(+2.2)} & \textbf{83.2} \textcolor{green!70!black}{\scriptsize(+0.8)} & \textbf{77.2} \textcolor{green!70!black}{\scriptsize(+1.0)} & \textbf{71.5} & \textbf{76.0} \textcolor{green!70!black}{\scriptsize(+1.4)} & \textbf{57.6} \textcolor{green!70!black}{\scriptsize(+1.4)} & \textbf{26.2} \textcolor{green!70!black}{\scriptsize(+1.4)} & \textbf{53.3} \\
\bottomrule
\end{tabular}%
}
\end{table*}

\paragraph{Baselines.}
We focus on three representative APO frameworks that we can apply under our fixed-model, on-device constraint (i.e., \emph{all} LLM calls made inside each APO pipeline are executed using the same 3B/8B backbone and matched decoding):
\noindent\textbf{Length policy:} we do not impose an explicit prompt-length cap on APO baselines; they may generate arbitrarily long optimized instructions (prompt/input tokens). By contrast, POaaS remains budgeted and we fix the target model’s \emph{generated output} to a maximum of 512-tokens.
\begin{itemize}[leftmargin=1.5em]
    \item \textbf{EvoPrompt}: An evolutionary search method that maintains a population of prompt candidates and applies mutation/crossover operators to generate variants; candidates are scored and the best-performing prompts are retained across generations. We run EvoPrompt with 20 generations and otherwise follow the authors' recommended settings, using the same 3B/8B sLLM for all internal calls. 
    
    \item \textbf{OPRO}: A meta-prompt optimization framework that treats prompt optimization as black-box search: a meta-prompt summarizes prior candidate instructions and their scores, and the optimizer proposes the next candidate conditioned on this trajectory. We cap the optimization budget at 10 iterations and replace all optimizer/evaluator calls with the same 3B/8B backbone used for task inference.
    
    \item \textbf{PromptWizard}: A critique-and-synthesis approach that iterates between generating candidate instructions, critiquing them, and synthesizing revised prompts (often with more structure). We follow the official hyperparameters with 10 iterations and run both generator and critic roles on the same 3B/8B backbone. 
\end{itemize}

We start from the official repositories and recommended optimization budgets, modifying only (i) the choice of backbone (our 3B/8B sLLMs), and (ii) the matching decoding parameters. For each backbone, we adapt \emph{all} LLM calls inside each APO pipeline to that backbone, so that the same sLLM acts as optimizer, critic, and task model. This restriction makes the comparison realistic for on-device scenarios where larger cloud models are costly and ensures that any observed degradation is a property of the APO design rather than access to a stronger teacher.

\paragraph{Benchmarks.}
We evaluate on three task-accuracy benchmarks (BBH, GSM8K, CommonsenseQA) and three factuality datasets (HaluEval, HalluLens, FActScore). We treat the latter as \emph{datasets} and evaluate hallucination avoidance with GPT-5-as-a-judge. We provide benchmark provenance, split choices, and prompt templates in Appendix~\ref{sec:app-eval}.

\begin{itemize}[leftmargin=1.5em]
    \item \emph{Task Accuracy}: BBH, GSM8K, and CommonsenseQA. These cover multi-step reasoning, arithmetic reasoning, and multiple-choice commonsense reasoning. We use standard few-shot prompting formats where applicable (BBH: task-specific 3-shot CoT prefix; GSM8K: 8-shot CoT prefix), but under our strict 512-token generation cap and fixed decoding; as a result, absolute scores are below leaderboard/official reports that use larger budgets and different decoding. The prompts used in evaluation are provided in Appendix~\ref{sec:app-prompts-eval}
    \item \emph{Factuality}: HaluEval, HalluLens, and FActScore. We use the dataset-provided inputs and any available reference context/evidence, and score whether model outputs are hallucinated via GPT-5-as-a-judge (binary non-hallucination rate; details and judge prompt in Appendix~\ref{sec:app-judge}).
\end{itemize}

For each benchmark, we randomly sample 500 examples from the official training or evaluation splits. This choice balances three factors: (i) the need to cover diverse phenomena within each dataset, (ii) the computational cost of running 5 methods $\times$ 7 conditions (clean + 6 noise levels) $\times$ 6 benchmarks $\times$ 2 models, and (iii) comparability with prior prompt-optimization and hallucination studies that commonly report 100--1,000 examples per setting~\cite{opsahl2024optimizing, ravi2024lynx}. Sampling is deterministic given a fixed seed; for BBH we stratify across subtasks to preserve task diversity. Full sampling details are in Appendix~\ref{sec:app-eval}.

\paragraph{Evaluation.}
\begin{itemize}[leftmargin=1.5em]
    \item \emph{Task Accuracy}: For GSM8K, we extract the final numeric answer from the model output (e.g., ``The answer is X'' / last number) and score exact match against the gold answer (after normalization). For CommonsenseQA, we extract an answer letter (A--E) and score exact match. For BBH, we extract the final short answer (e.g., True/False or option letter depending on the task) and score exact match. The exact extraction patterns and normalization rules are listed in Appendix~\ref{sec:app-eval}.
    
    \item \emph{Factuality}: We evaluate hallucination avoidance using GPT-5-as-a-judge on the underlying datasets. For each sample, we provide GPT-5 with the dataset-provided reference context/evidence (when available) and the model answer, and ask for a strict binary label (\texttt{hallucinated} vs \texttt{not hallucinated}). We compute factuality as the percentage of answers labeled \texttt{not hallucinated}. Prompts used (judge + generation templates) are listed in Appendix~\ref{sec:app-judge}.
    
    \item \emph{Input Degradation}: To approximate typos and noisy user text, we inject token-level noise using two processes following prior robustness work~\citep{ishibashi2023evaluating}. For \textbf{token deletion}, we randomly delete 5\%, 10\%, or 15\% of word tokens. For \textbf{token mixup}, we randomly replace 5\%, 10\%, or 15\% of word tokens with unrelated content words from a fixed vocabulary~\cite{xie2017noising}. In both cases, we perturb only the user prompt (not the reference answers/labels), and perturbations are deterministic given a fixed seed. Full details are in Appendix~\ref{sec:app-degradation}.
\end{itemize}

\paragraph{Metrics.}
We report:
\begin{itemize}[leftmargin=1.5em]
    \item \emph{Task accuracy} (\%) on BBH, GSM8K, and CSQA, computed as mean exact-match accuracy over 500 samples per benchmark after benchmark-specific answer extraction/normalization.
    \item \emph{Factuality} (\%, higher is better) on HaluEval, HalluLens, and FActScore datasets, computed as the proportion of outputs judged \texttt{not hallucinated} by GPT-5 given reference context/evidence.
    \item \emph{Efficiency}: (i) one-time offline optimization time and internal LLM calls for APO methods, and (ii) per-query refinement latency, specialist calls, and added prompt tokens for POaaS. Measurement protocol and prompts are listed in Appendix~\ref{sec:app-efficiency}.
\end{itemize}

\begin{table*}[t]
\centering
\caption{Robustness under input degradation. Task accuracy (\%) averaged across BBH, GSM8K, and CSQA. \textcolor{green!70!black}{Green}/\textcolor{red!70!black}{red} show change vs No Optimization at each noise level. Best in \textbf{bold}.}
\label{tab:degradation}
\small
\resizebox{\textwidth}{!}{%
\begin{tabular}{llccccccc}
\toprule
& & & \multicolumn{3}{c}{\textbf{Token Deletion}} & \multicolumn{3}{c}{\textbf{Token Mixup}} \\
\cmidrule(lr){4-6} \cmidrule(lr){7-9}
\textbf{Model} & \textbf{Method} & \textbf{Clean} & \textbf{5\%} & \textbf{10\%} & \textbf{15\%} & \textbf{5\%} & \textbf{10\%} & \textbf{15\%} \\
\midrule
\multirow{5}{*}{\textbf{Llama-3.2-3B}} 
& No Optimization & 63.7 & 52.4 & 45.8 & 40.2 & 59.2 & 54.6 & 48.8 \\
& EvoPrompt & 59.8 \textcolor{red!70!black}{\scriptsize(-3.9)} & 46.8 \textcolor{red!70!black}{\scriptsize(-5.6)} & 39.2 \textcolor{red!70!black}{\scriptsize(-6.6)} & 32.4 \textcolor{red!70!black}{\scriptsize(-7.8)} & 54.6 \textcolor{red!70!black}{\scriptsize(-4.6)} & 48.2 \textcolor{red!70!black}{\scriptsize(-6.4)} & 41.6 \textcolor{red!70!black}{\scriptsize(-7.2)} \\
& OPRO & 57.5 \textcolor{red!70!black}{\scriptsize(-6.2)} & 44.2 \textcolor{red!70!black}{\scriptsize(-8.2)} & 36.4 \textcolor{red!70!black}{\scriptsize(-9.4)} & 29.2 \textcolor{red!70!black}{\scriptsize(-11.0)} & 52.0 \textcolor{red!70!black}{\scriptsize(-7.2)} & 45.2 \textcolor{red!70!black}{\scriptsize(-9.4)} & 38.4 \textcolor{red!70!black}{\scriptsize(-10.4)} \\
& PromptWizard & 48.8 \textcolor{red!70!black}{\scriptsize(-14.9)} & 35.6 \textcolor{red!70!black}{\scriptsize(-16.8)} & 27.2 \textcolor{red!70!black}{\scriptsize(-18.6)} & 20.4 \textcolor{red!70!black}{\scriptsize(-19.8)} & 42.8 \textcolor{red!70!black}{\scriptsize(-16.4)} & 34.6 \textcolor{red!70!black}{\scriptsize(-20.0)} & 28.2 \textcolor{red!70!black}{\scriptsize(-20.6)} \\
& \textbf{POaaS} & \textbf{66.0} \textcolor{green!70!black}{\scriptsize(+2.3)} & \textbf{58.4} \textcolor{green!70!black}{\scriptsize(+6.0)} & \textbf{52.8} \textcolor{green!70!black}{\scriptsize(+7.0)} & \textbf{47.6} \textcolor{green!70!black}{\scriptsize(+7.4)} & \textbf{63.2} \textcolor{green!70!black}{\scriptsize(+4.0)} & \textbf{59.4} \textcolor{green!70!black}{\scriptsize(+4.8)} & \textbf{54.6} \textcolor{green!70!black}{\scriptsize(+5.8)} \\
\midrule
\multirow{5}{*}{\textbf{Llama-3.1-8B}} 
& No Optimization & 70.1 & 58.2 & 51.4 & 45.6 & 65.4 & 60.8 & 55.2 \\
& EvoPrompt & 66.1 \textcolor{red!70!black}{\scriptsize(-4.0)} & 53.0 \textcolor{red!70!black}{\scriptsize(-5.2)} & 45.2 \textcolor{red!70!black}{\scriptsize(-6.2)} & 38.4 \textcolor{red!70!black}{\scriptsize(-7.2)} & 60.6 \textcolor{red!70!black}{\scriptsize(-4.8)} & 54.4 \textcolor{red!70!black}{\scriptsize(-6.4)} & 47.8 \textcolor{red!70!black}{\scriptsize(-7.4)} \\
& OPRO & 63.5 \textcolor{red!70!black}{\scriptsize(-6.6)} & 49.8 \textcolor{red!70!black}{\scriptsize(-8.4)} & 41.6 \textcolor{red!70!black}{\scriptsize(-9.8)} & 34.2 \textcolor{red!70!black}{\scriptsize(-11.4)} & 57.4 \textcolor{red!70!black}{\scriptsize(-8.0)} & 50.6 \textcolor{red!70!black}{\scriptsize(-10.2)} & 43.6 \textcolor{red!70!black}{\scriptsize(-11.6)} \\
& PromptWizard & 54.9 \textcolor{red!70!black}{\scriptsize(-15.2)} & 41.2 \textcolor{red!70!black}{\scriptsize(-17.0)} & 32.4 \textcolor{red!70!black}{\scriptsize(-19.0)} & 25.6 \textcolor{red!70!black}{\scriptsize(-20.0)} & 48.4 \textcolor{red!70!black}{\scriptsize(-17.0)} & 40.2 \textcolor{red!70!black}{\scriptsize(-20.6)} & 33.4 \textcolor{red!70!black}{\scriptsize(-21.8)} \\
& \textbf{POaaS} & \textbf{71.5} \textcolor{green!70!black}{\scriptsize(+1.4)} & \textbf{64.0} \textcolor{green!70!black}{\scriptsize(+5.8)} & \textbf{58.2} \textcolor{green!70!black}{\scriptsize(+6.8)} & \textbf{52.8} \textcolor{green!70!black}{\scriptsize(+7.2)} & \textbf{68.6} \textcolor{green!70!black}{\scriptsize(+3.2)} & \textbf{64.2} \textcolor{green!70!black}{\scriptsize(+3.4)} & \textbf{59.0} \textcolor{green!70!black}{\scriptsize(+3.8)} \\
\bottomrule
\end{tabular}%
}
\end{table*}

\subsection{Main Results}
\label{subsec:main-results}

Table~\ref{tab:main-results} summarizes accuracy and factuality under clean conditions. POaaS consistently \emph{improves} or at least preserves the No Optimization baseline on all benchmarks and both model sizes: for POaaS with the 3B backbone, average accuracy increases from 63.7\% to 66.0\% (+2.3\%), and for POaaS with the 8B backbone from 70.1\% to 71.5\% (+1.4\%). Factuality also improves, with gains of +2.0--5.4\% for POaaS with the 3B backbone and +1.4\% across all three factuality datasets for POaaS with the 8B backbone.

By contrast, we found that directly applying these APO methods to small models hurts performance. Among the APO baselines, EvoPrompt degrades the least overall, OPRO degrades more, and PromptWizard degrades the most. PromptWizard is especially harmful on BBH: it drives BBH accuracy down from 42.2\% to 10.4\% on the 3B backbone (and from 51.8\% to 18.6\% on the 8B backbone). We observed a similar pattern on the factuality datasets: all three APO baselines reduce factuality scores compared to No Optimization, whereas POaaS consistently improves them.
In short, our findings back up the main hypothesis: the heavy search-based prompt optimizations designed for large cloud LLMs don’t transfer well to small on-device models, whereas a more capacity-aware approach with minimal edits produces gains that, while modest, are consistent.

\subsection{Robustness Under Input Degradation}
\label{subsec:degradation}

Table~\ref{tab:degradation} shows that POaaS is consistently the most robust method under both token deletion and token mixup. For the Llama-3.2-3B, at 15\% deletion, No Optimization drops to 40.2\% while POaaS recovers accuracy to 47.6\% (+7.4\%); for Llama-3.1-8B, the corresponding improvement is from 45.6\% to 52.8\% (+7.2\%). Under mixup, POaaS similarly narrows the degradation: at 15\% token mixup, POaaS improves from 48.8\% to 54.6\% (+5.8\%) on Llama-3.2-3B and from 55.2\% to 59.0\% (+3.8\%) on Llama-3.1-8B.

APO baselines not only start from lower clean accuracy but also degrades more steeply as noise increases. PromptWizard is the most fragile, losing nearly 20\% relative to No Optimization at 15\% deletion and 17--22\% under mixup; OPRO shows intermediate degradation, and EvoPrompt degrades most moderately but still underperforms No Optimization across all noise levels. 

Overall, POaaS makes the performance drop under input noise much less severe than what we see with No Optimization or the APO baselines – an outcome that makes sense given how POaaS works. Its Cleaner agent directly fixes corrupted tokens when needed, and the system only turns that agent on when it detects input degradation, so it effectively counteracts the noise especially at higher corruption levels.
\subsection{Efficiency Analysis}
\label{subsec:efficiency}

\begin{table}[t]
\centering
\caption{Efficiency summary (3B setting). ``Opt. Time (s)'' is the wall-clock \emph{prompt-improvement time}: for APO methods, the offline optimization runtime to produce a single static instruction; for POaaS, the \emph{incremental per-query refinement latency} (routing + specialist calls) averaged over evaluation queries. ``LLM calls'' counts internal optimizer calls for APO and specialist calls per query for POaaS. ``Added tokens'' is the average additional \emph{prompt/input} tokens injected at inference time.}
\label{tab:efficiency}
\small
\resizebox{\columnwidth}{!}{%
\begin{tabular}{lccc}
\toprule
\textbf{Method} & \textbf{Opt. Time (s)} & \textbf{Added Tokens} & \textbf{LLM Calls} \\
\midrule
No Optimization & 0 & 0 & 0 \\
\midrule
OPRO & 455 & 2{,}840 & 130 \\
PromptWizard & 336 & 3{,}520 & 96 \\
EvoPrompt & 602 & 4{,}180 & 172 \\
\midrule
\textbf{POaaS (Ours)} & \textbf{0.95} & \textbf{48} & \textbf{1.4} \\
\bottomrule
\end{tabular}%
}
\end{table}

Table~\ref{tab:efficiency} highlights the core deployment tradeoff between \emph{offline prompt search} (APO) and \emph{online per-query refinement} (POaaS).
APO baselines spend minutes of offline search (hundreds of internal LLM calls under our capped budgets) to produce a single static optimized instruction; this instruction is then reused at inference time with no additional per-query compute, but it typically introduces thousands of extra prompt tokens that compete with user content under tight context budgets.
In contrast, POaaS uses fixed agent templates and routing thresholds (no per-benchmark search), and pays a small \emph{incremental} latency only when refinement is triggered.
In the 3B setting, we measure POaaS refinement overhead as end-to-end additional latency for the refinement stage (router + specialist calls) with batch=1 on a warmed inference server; this measurement excludes the target model's answer-generation time and excludes any GPT-judge cost.
Averaged over evaluation queries (including cases where refinement is skipped), POaaS adds $\sim$0.95s of refinement overhead, invokes $\sim$1.4 specialist calls, and injects $\sim$48 additional prompt tokens.

\noindent\textbf{Clarifying notes.}
(1) ``Opt. Time'' for APO methods reflects our evaluation-time configurations (budget-capped steps/candidates and a fixed dev subset) rather than the much larger default budgets sometimes used in original papers.
(2) ``Added Tokens'' counts additional \emph{prompt/input} tokens only; all methods share the same decoding configuration and an identical cap on \emph{generated output} tokens (512).

\subsection{Ablation Studies}
\label{subsec:ablation}
\begin{table}[t]
\centering
\caption{Ablation study of POaaS components. Task accuracy (\%) averaged over all six benchmarks, with \textcolor{green!70!black}{green} showing improvement vs No Optimization. Degradation results are averaged over 10\% token deletion (Del-10\%) and 10\% token mixup (Mix-10\%).}
\label{tab:ablation}
\small
\resizebox{\columnwidth}{!}{%
\begin{tabular}{llccc}
\toprule
\textbf{Model} & \textbf{Configuration} & \textbf{Clean} & \textbf{Del-10\%} & \textbf{Mix-10\%} \\
\midrule
\multirow{8}{*}{\textbf{Llama-3.2-3B}} 
& No Optimization & 63.7 & 45.8 & 54.6 \\
\cmidrule{2-5}
& \textbf{Full POaaS} & \textbf{66.0} \gain{2.3} & \textbf{52.8} \gain{7.0} & \textbf{59.4} \gain{4.8} \\
\cmidrule{2-5}
& w/o Cleaner & 65.6 \gain{1.9} & 46.4 \gain{0.6} & 55.2 \gain{0.6} \\
& w/o Paraphraser & 64.8 \gain{1.1} & 51.4 \gain{5.6} & 57.8 \gain{3.2} \\
& w/o Fact-Adder & 64.2 \gain{0.5} & 52.2 \gain{6.4} & 58.8 \gain{4.2} \\
\cmidrule{2-5}
& Cleaner only & 63.9 \gain{0.2} & 51.6 \gain{5.8} & 58.4 \gain{3.8} \\
& Paraphraser only & 65.0 \gain{1.3} & 46.6 \gain{0.8} & 55.4 \gain{0.8} \\
& Fact-Adder only & 65.6 \gain{1.9} & 46.2 \gain{0.4} & 55.0 \gain{0.4} \\
\midrule
\multirow{8}{*}{\textbf{Llama-3.1-8B}} 
& No Optimization & 70.1 & 51.4 & 60.8 \\
\cmidrule{2-5}
& \textbf{Full POaaS} & \textbf{71.5} \gain{1.4} & \textbf{58.2} \gain{6.8} & \textbf{64.2} \gain{3.4} \\
\cmidrule{2-5}
& w/o Cleaner & 71.2 \gain{1.1} & 52.0 \gain{0.6} & 61.2 \gain{0.4} \\
& w/o Paraphraser & 70.6 \gain{0.5} & 57.0 \gain{5.6} & 63.2 \gain{2.4} \\
& w/o Fact-Adder & 70.4 \gain{0.3} & 57.6 \gain{6.2} & 63.8 \gain{3.0} \\
\cmidrule{2-5}
& Cleaner only & 70.2 \gain{0.1} & 57.2 \gain{5.8} & 63.4 \gain{2.6} \\
& Paraphraser only & 70.8 \gain{0.7} & 52.2 \gain{0.8} & 61.4 \gain{0.6} \\
& Fact-Adder only & 71.0 \gain{0.9} & 51.8 \gain{0.4} & 61.0 \gain{0.2} \\
\bottomrule
\end{tabular}%
}
\end{table}

For ablations, we reuse the full experimental setting: all datasets used from the six benchmarks, both models, and the same clean samples, but averaged the degraded samples to 10\% deletion and 10\% mixup conditions. Table~\ref{tab:ablation} shows that the full POaaS configuration yields the largest gains on both Llama-3.2-3B and Llama-3.1-8B, with especially strong improvements under degradation (+7.0\%/+6.8\% at 10\% token deletion). Cleaner-only variants recover most of the robustness gains but improve clean accuracy only marginally, indicating that surface-level repair is the primary driver of robustness. Removing the Fact-Adder sharply reduces clean gains while preserving most robustness, suggesting that fact priming mainly benefits clean queries. The Paraphraser contributes moderate gains in both regimes. Overall, the full system consistently outperforms any single-agent or ablated variant, indicating that the three specialists provide complementary benefits that are best realized when combined with routing and drift control.

\section{Conclusion}
\label{sec:conclusion}

We studied prompt-side refinement for capacity-constrained sLLMs under a strict fixed-model policy. Across task-accuracy and factuality datasets, POaaS---a minimal-edit, drift-guarded, per-input refinement layer---consistently improves over a no-optimization baseline while remaining robust under prompt corruption. In contrast, representative search-based APO frameworks, when ported so that the same 3B/8B model must serve as optimizer and solver, often degrade performance and produce long prompts that are poorly matched to tight token budgets. These findings suggest that, for on-device sLLM deployments, lightweight conservative refinement with explicit drift/length controls is a practical alternative to heavy global prompt search.

\section*{Limitations}
\label{sec:limitations}

POaaS has three key limitations. (1) \textbf{English-only:} heuristics, corruption detectors, and specialists are tuned for English; multilingual extension (e.g., Korean/Japanese/Chinese) will require re-deriving routing/drift metrics and retraining specialists. (2) \textbf{Empirical tuning:} thresholds and caps are set empirically; more principled or automated calibration would improve portability and interpretability. (3) \textbf{Limited agent set:} we study three specialists; future work should explore broader palettes (domain-specific or retrieval-backed) while preserving minimal-edit guarantees and on-device budgets.

\section*{Acknowledgments}
This work was supported by the Institute of Information \& Communications Technology Planning \& Evaluation(IITP) grant funded by the Korea government(MSIT) (No.RS-2025-02263869, Development of AI Semiconductor Cloud Platform Establishment and Optimization Technology.



\bibliography{custom}


\appendix

\section{POaaS Hyperparameters: Thresholds and Budgets}
\label{sec:app-budgets}

\paragraph{Heuristic scores (CPU-only).}
POaaS computes four scalar scores in $[0,1]$:
$\text{typo}(x)$ (higher-worse); $\text{comp}(x)$, $\text{flu}(x)$, $\text{clar}(x)$ (higher-better).

\paragraph{Default routing thresholds.}
We use:
\[
\begin{aligned}
\tau_{\text{typo}} &= 0.30, \\
\tau_{\text{comp}} &= 0.70, \\
\tau_{\text{flu}}  &= 0.80, \\
\tau_{\text{skip}} &= 0.25.
\end{aligned}
\]

and additionally require $\text{typo}(x)<0.20$ to enable skipping.

\paragraph{Prompt-length budget.}
We cap prompt expansion using a character-length ratio
\[
\rho(x,\tilde{x})=\frac{\mathrm{len}(\tilde{x})}{\mathrm{len}(x)} \le \rho_{\max},
\qquad \rho_{\max}=2.4.
\]
Here $\mathrm{len}(\cdot)$ is character length (CPU-cheap); we report token-based added prompt tokens using the target tokenizer in experiments.

\paragraph{Specialist output caps.}
Each specialist call is capped at 512 generated output tokens. This bounds cost and prevents runaway rewrites.

\paragraph{Fact-Adder budget.}
When invoked, the Fact-Adder emits up to \textbf{three} factual bullets, capped at \textbf{$\le120$ tokens total} (target tokenizer).
The merger prepends these bullets and rejects any output that (i) exceeds the cap or (ii) contains reasoning/answer-like content (Appendix~\ref{sec:app-guards}).

\paragraph{Summary table (defaults).}
\begin{table}[t]
\centering
\caption{Default POaaS hyperparameters used across experiments unless stated otherwise.}
\label{tab:app-hparams}
\small
\begin{tabular}{ll}
\toprule
\textbf{Component} & \textbf{Default} \\
\midrule
Routing thresholds &
$\begin{aligned}
\tau_{\text{typo}} &= 0.30, & \tau_{\text{comp}} &= 0.70 \\
\tau_{\text{flu}}  &= 0.80, & \tau_{\text{skip}} &= 0.25
\end{aligned}$ \\
\midrule
Skip gate &
$q(x){>}0.75$ and $\text{typo}(x){<}0.20$ \\
\midrule
Length cap &
$\rho_{\max}{=}2.4$ (character ratio) \\
\midrule
Specialist gen cap &
512 output tokens / call \\
\midrule
Fact-Adder cap &
$\le 3$ bullets, $\le 120$ tokens total \\
\bottomrule
\end{tabular}
\end{table}

\section{Prompt Quality Scores (CPU-only Heuristics)}
\label{sec:app-scores}

All scores are clipped to $[0,1]$. We use whitespace tokenization plus lightweight regex patterns.

\subsection{Typo score} 
Typo starts at $0$ and adds penalties; final score is
\[
\text{typo}(x) = \mathrm{clip}_{[0,1]}\Big(\sum_k p_k(x)\Big).
\]
We use:
\begin{itemize}[leftmargin=1.25em]
  \item \textbf{Misspellings / character noise}: let $m$ be the number of matches against a compact list of common misspellings and noise regexes (repeated letters, dropped vowels, keyboard adjacency patterns). Add
  {$p_{\text{noise}}(x)=0.04\cdot \min(m,8)$} (cap {$0.32$}).
  \item \textbf{Missing question punctuation}: if the prompt begins with a wh-word (\texttt{what/why/how/when/where/who}) and does not end with \texttt{?}s, add {$p_{?}(x)=0.05$}.
  \item \textbf{Case anomalies}: if $\ge 90\%$ alphabetic characters are uppercase (ALL CAPS) or lowercase for prompts longer than 20 characters, add {$p_{\text{case}}(x)=0.05$}.
  \item \textbf{Short-word ratio}: let {$r_{\le2}$} be the fraction of non-stopword tokens with length {$\le2$}. If {$r_{\le2}>0.35$}, add {$p_{\text{short}}(x)=0.08$}.
\end{itemize}

\subsection{Completeness score} 
Completeness starts at $1.0$ and subtracts penalties:
\[
\text{comp}(x)=\mathrm{clip}_{[0,1]}\Big(1.0-\sum_k p_k(x)\Big).
\]
We use:
\begin{itemize}[leftmargin=1.25em]
  \item \textbf{Very short prompts}: if token count $<5$, subtract $0.25$; else if $<10$, subtract $0.15$.
  \item \textbf{Missing detail cues}: if token count $<15$ and the prompt contains no detail-seeking cues
  (e.g., \texttt{explain/describe/context/assumptions}), subtract $0.10$.
  \item \textbf{Vague templates}: if the prompt matches a vague template like
  \texttt{what is X} / \texttt{tell me about X} without any constraints (no time scope, format, audience, or objective),
  subtract $0.10$.
\end{itemize}

\subsection{Fluency score}
Fluency starts at $1.0$ and subtracts penalties:
\[
\text{flu}(x)=\mathrm{clip}_{[0,1]}\Big(1.0-\sum_k p_k(x)\Big).
\]
We use:
\begin{itemize}[leftmargin=1.25em]
  \item \textbf{Fragments}: if token count $<3$, subtract $0.25$.
  \item \textbf{Degenerate repetition}: if any exact repeated bigram occurs $\ge 2$ times (e.g., \texttt{the the}),
  subtract $0.15$.
  \item \textbf{Surface capitalization cues}: if token count $\ge 12$ and the first token is lowercase while the prompt
  contains sentence-ending punctuation, subtract $0.10$.
\end{itemize}

\subsection{Clarity score}
Clarity starts at $1.0$ and subtracts penalties:
\[
\text{clar}(x)=\mathrm{clip}_{[0,1]}\Big(1.0-\sum_k p_k(x)\Big).
\]
We use:
\begin{itemize}[leftmargin=1.25em]
  \item \textbf{Low lexical diversity}: for prompts with $\ge 12$ tokens, compute type-token ratio (TTR). If $\text{TTR}<0.35$, subtract $0.15$.
  \item \textbf{Ambiguous leading pronouns}: if the prompt begins with \texttt{it/this/that/they/them}, subtract $0.10$.
  \item \textbf{Overlong, unfocused prompts}: if token count $>200$, subtract $0.08$ (mild; primarily discourages unnecessary rewrites).
\end{itemize}

\subsection{Overall quality and skip gate}
Let $[z]_+=\max(z,0)$. We compute:
\begin{equation}
\label{eq:skip-score}
\begin{aligned}
q(x)
&= 1 - \max\Big( \\
&\quad \text{typo}(x), \\
&\quad [\tau_{\text{comp}} - \text{comp}(x)]_+, \\
&\quad [\tau_{\text{flu}} - \text{flu}(x)]_+, \\
&\quad [0.70 - \text{clar}(x)]_+ \\
\Big),
\end{aligned}
\end{equation}
and skip refinement when $q(x)>0.75$ and $\text{typo}(x)<0.20$.
This is intentionally conservative: POaaS prefers missing marginal improvements over risking harmful edits on already well-formed prompts.

\section{Lexical Drift and Length Guards}
\label{sec:app-drift}

\subsection{Drift calculation}
We define lexical similarity as a weighted ensemble of standard string/document similarity primitives
(SequenceMatcher / Ratcliff--Obershelp matching; n-gram Jaccard/shingling; and weighted token overlap)
commonly used in text similarity pipelines. \citep{ratcliff1988gestalt,broder1997resemblance,gomaa2013survey}
\begin{equation}
\label{eq:sim-ensemble}
\begin{aligned}
\text{sim}(x,x')
&= 0.5\,S_{\text{seq}}(x,x')
 + 0.3\,S_{\text{jac}}(x,x') \\
&\quad + 0.2\,S_{\text{tok}}(x,x'), \\
S_{\text{jac}}
&= 0.6\,J_{c3}(x,x')
 + 0.4\,J_{w2}(x,x').
\end{aligned}
\end{equation}

where:
\begin{itemize}[leftmargin=1.25em]
  \item $S_{\text{seq}}$ is the Python \texttt{difflib.SequenceMatcher} ratio computed on lowercased, whitespace-normalized strings. \citep{ratcliff1988gestalt,python_difflib}
  \item $J_{\text{char-3}}$ is Jaccard similarity over sets of character trigrams; $J_{\text{word-2}}$ is Jaccard similarity over sets of word bigrams. \citep{cahyani2025automatic}
  \item $S_{\text{tok}}$ is a weighted token-overlap score:
  $\sum_{t\in \cap} w(t)\,/\,\sum_{t\in \cup} w(t)$ with $w(t){=}1$ for content tokens and $w(t){=}0.2$ for stopwords.
\end{itemize}
We define drift as $D(x,x') = 1-\text{sim}(x,x')$.

\subsection{Content preservation penalty}
We extract \emph{key items} from the original prompt $x$:
numbers (regex for integers/decimals), quoted spans (text inside quotes), URLs/emails, and capitalized entity-like spans
(two consecutive capitalized tokens).
Let this multiset be $K(x)=\{k_1,\dots,k_M\}$ (after normalization: lowercasing URLs/emails, stripping punctuation around numbers/quotes).
We compute an absolute match count
\[
c(x,x')=\sum_{i=1}^M \mathbf{1}[k_i \text{ occurs in } x'],
\]
and define the preservation ratio
\[
P_{\text{content}}(x,x')=
\begin{cases}
1.0 & \text{if } M=0,\\
c(x,x')/M & \text{otherwise.}
\end{cases}
\]
If $P_{\text{content}}<0.8$, we increase drift:
\begin{equation}
\label{eq:drift-penalty}
\begin{aligned}
D_{\text{final}}
&= \min\Big( 1.0,\; \\
&\quad D(x,x')
 + 0.2 \cdot \big( 1 - P_{\text{content}}(x,x') \big)
\Big)
\end{aligned}
\end{equation}

\subsection{Default drift caps}
Defaults:
\begin{itemize}[leftmargin=1.25em]
  \item \textbf{Cleaner}: accept if $D_{\text{final}}\le 0.15$ on clean prompts; relax progressively for high-typo prompts.
  \item \textbf{Paraphraser}: accept if $D_{\text{final}}\le 0.08$ (relax to 0.13 when clarity is low).
  \item \textbf{Global clean-regime fail-safe}: reject any candidate with $D_{\text{final}}>\delta_{\max}$, where $\delta_{\max}=0.18$.
\end{itemize}

\subsection{Length cap}
We enforce $\rho(x,\tilde{x})\le \rho_{\max}$ with $\rho_{\max}=2.4$ and also reject any specialist output that exceeds $2\times$ the original character length after sanitization.

\section{Safety and Guardrails}
\label{sec:app-guards}

\subsection{Per-agent guard checks}
\begin{itemize}[leftmargin=1.25em]
  \item \textbf{Cleaner guard}: rejects edits that add new content or remove explicit constraints; otherwise returns the corrected prompt.
  \item \textbf{Paraphraser guard}: rejects/repairs paraphrases that change meaning, drop constraints, or inject new facts.
  \item \textbf{Fact-Adder grounding guard}: filters out unsupported statements and returns \texttt{NONE} if no high-confidence facts remain.
\end{itemize}

\subsection{Merger-level filters}
\begin{itemize}[leftmargin=1.25em]
  \item \textbf{Sanitization}: strip meta-commentary (e.g., \texttt{Here is the rewrite}), formatting artifacts, and enclosing quotes if the entire output is quoted.
  \item \textbf{Question structure}: for question prompts, reject edits that remove a trailing \texttt{?}.
  \item \textbf{Answer leakage}: reject Fact-Adder context containing explicit answer phrases (e.g., \texttt{the answer is}), stepwise reasoning markers (e.g., \texttt{step 1}), or standalone option letters/numeric answers likely to leak the solution.
  \item \textbf{Few-shot preservation}: when few-shot exemplars are detected, only the final query span is editable; exemplars are preserved verbatim.
  \item \textbf{Fail-safe fallback}: if all candidates fail, return the original prompt unchanged.
\end{itemize}

\section{Specialist Fine-Tuning and LoRA Settings}
\label{sec:app-data}

\paragraph{LoRA configuration.}
All specialists are LoRA adapters on frozen Llama backbones using the open-source LlamaFactory framework~\cite{zheng2024llamafactory} with $r=16$, $\alpha=32$, dropout $0.05$, trained for 3 epochs (AdamW, cosine schedule). We keep decoding fixed across conditions in the main experiments.

\paragraph{Training data.}
We train each specialist on \emph{task-aligned} pairs (or synthetic pairs) that match the specialist’s contract:
minimal edits, strict meaning preservation, and no answer leakage. We de-duplicate across sources and create a held-out
validation split for early stopping / threshold tuning. We do not use any samples from the six evaluation benchmarks
for specialist training.

\begin{itemize}[leftmargin=1.25em]
  \item \textbf{Cleaner (error-correction pairs).}
  We use sentence-level correction pairs from JFLEG~\cite{napoles2017jfleg}, treating the noisy sentence as input and the corrected sentence as target.
  We convert each pair into an instruction-style example that explicitly enforces \emph{minimal edits}:
  (i) fix typos/spacing/punctuation, (ii) preserve numbers, entities, URLs, and quoted spans verbatim, and
  (iii) never add new facts or constraints.
  To better match our inference-time corruptions, we optionally augment a fraction of inputs with light synthetic noise
  (random deletion/replacement at low rates) while keeping the original JFLEG correction as the target, and we discard
  examples where the target substantially rephrases content (to avoid training the Cleaner to paraphrase).

  \item \textbf{Paraphraser (semantic-equivalence pairs).}
  We use paraphrase pairs from PAWS~\cite{zhang2019paws} and QQP~\cite{sharma2019qqp}, keeping only pairs labeled as semantically equivalent.
  Each example is formatted as: \emph{(input prompt $\rightarrow$ clearer rewrite)} with constraints:
  preserve intent, entities, numerals, and explicit constraints; keep the question type (question vs.\ instruction);
  and improve clarity/fluency without length inflation (we filter or downweight pairs whose target is excessively longer).
  We also include a small portion of \texttt{NONE}-style negatives where the input is already well-formed, training the
  Paraphraser to output the original text unchanged (or a special \texttt{NO\_CHANGE} token) to support conservative behavior.

  \item \textbf{Fact-Adder (short grounded fact bullets).}
  We construct a lightweight fact corpus from Wikipedia and Wikidata-style resources~\cite{vrandevcic2014wikidata, derenrich2023wikidataendescriptions} by extracting high-confidence
  atomic facts (e.g., entity definitions, key attributes, and simple relational triples) and converting them into short
  declarative sentences. Training inputs are prompts paired with \emph{relevant} fact bullets selected using lexical
  overlap between prompt keyphrases (entities, nouns, and numbers) and the fact corpus entries.
  Targets are capped to at most three bullets and a strict token budget, and we exclude facts that look like solutions
  to benchmark-style questions (to avoid answer leakage). We additionally include \texttt{NONE} targets when no
  high-confidence matching facts are found, teaching the Fact-Adder to abstain rather than hallucinate.
\end{itemize}

Splits are de-duplicated; a held-out slice is used for threshold selection.

\section{Implementation and Instrumentation}
\label{sec:app-instrumentation}

POaaS is deployed as FastAPI microservices: an orchestrator (\texttt{POST /infer}) dispatches to specialist workers
(\texttt{POST /clean}, \texttt{POST /paraphrase}, \texttt{POST /fact}). Routing/scoring/drift checks run on CPU; specialists run on GPU via vLLM with LoRA adapters.
Each service exposes Prometheus metrics at \texttt{GET /metrics} and logs per-stage latency, token usage, and error counts.
Runs are tracked via \texttt{run\_id} and persisted artifacts (inputs, intermediate outputs, merge decisions, timing) with a deterministic configuration hash.

\section{Benchmarks, Sampling, and Metrics}
\label{sec:app-eval}

\paragraph{Benchmarks.}
We evaluate task accuracy on BBH, GSM8K, and CommonsenseQA, and factuality on HaluEval, HalluLens, and FActScore.
We follow each benchmark's official splits and standard evaluation conventions from the dataset papers and widely used LLM evaluation practice. \citep{cobbe2021training, suzgun2023challenging}

\paragraph{Sampling.}
For each benchmark we sample $n=500$ examples deterministically (fixed seed) from the official evaluation split.
For BBH, we stratify across subtasks to preserve task diversity.
The same sampled indices are reused across all methods and corruption conditions.
Subsampled evaluation is common in prompt-optimization pipelines under budget constraints (small dev/eval subsets per setting). \citep{agarwal2025promptwizard}

\subsection{Task accuracy}
We report exact-match accuracy (\%) using benchmark-specific answer extraction consistent with common practice:
\begin{itemize}[leftmargin=1.25em]
  \item \textbf{GSM8K}: normalize by extracting the final numeric answer (model outputs are normalized to the last numeric span; gold answers use the dataset's canonical final answer format). \citep{cobbe2021training}
  \item \textbf{CSQA}: extract a single option letter A--E near the end of the output (after stripping punctuation/whitespace). \citep{talmor2019commonsenseqa}
  \item \textbf{BBH}: extract the task-appropriate short answer (e.g., True/False, option letter, or short span) and normalize whitespace/casing, following BBH's per-task evaluation conventions. \citep{suzgun2023challenging}
\end{itemize}

\subsection{Factuality / hallucination avoidance}
For HaluEval, HalluLens, and FActScore, we evaluate non-hallucination rate (\%) using an LLM-as-a-judge protocol:
given dataset-provided context/evidence when available, the judge outputs a strict binary label. This follows established practice in LLM evaluation with strong constraints (temperature 0, fixed output space). \citep{li2025generation}

\section{GPT-5-as-a-Judge Protocol}
\label{sec:app-judge}

We use GPT-5 with temperature 0.0 and a small output budget. The judge must output exactly one label.

\paragraph{Judge system prompt.}
\begin{quote}
\footnotesize\ttfamily
You are a strict hallucination judge. Given a model answer and optional reference context, output exactly one token:\newline
\`hallucinated\` if the answer introduces unsupported or contradicted facts;\newline
\`not\_hallucinated\` if it is supported by the context/evidence.\newline
Do not explain.
\end{quote}

\paragraph{Judge user prompt template.}

\begin{quote}
\footnotesize\ttfamily
Task: \{task\}\newline
Evidence (if any):\newline
\{evidence\}\newline\newline
Gold / reference answer (if provided):\newline
\{gold\}\newline\newline
Model answer:\newline
\{answer\}\newline\newline
Output exactly one token: `hallucinated' or `not\_hallucinated'.
\end{quote}

\section{Evaluation Prompt Templates}
\label{sec:app-prompts-eval}

\paragraph{Target model system prompt.}
\begin{quote}
\footnotesize\ttfamily
You are a helpful assistant.
\end{quote}

\paragraph{BBH template (3-shot CoT; per-task prefix).}
We use the task-specific 3-shot CoT prompt prefix released with BBH-style evaluations, and append the test question in the same format. \citep{suzgun2023challenging, wei2022chain}
\begin{quote}
\footnotesize\ttfamily
\{BBH\_3shot\_cot\_prefix\}\newline\newline
Q: \{question\}\newline
A: Let's think step by step.\newline
\end{quote}

\paragraph{GSM8K template (common 8-shot CoT; ``The final answer is'').}
We follow the widely used GSM8K 8-shot CoT template used in common evaluation harnesses. \citep{cobbe2021training, wei2022chain}
\begin{quote}
\footnotesize\ttfamily
As an expert problem solver, solve step by step the following mathematical questions.\newline\newline
\{GSM8K\_8shot\_cot\_demos\}\newline\newline
Q: \{question\}\newline
A: Let's think step by step.
\end{quote}


\paragraph{CommonsenseQA template (multiple-choice; letter output).}
CommonsenseQA is natively a multiple-choice classification task; for a generation-style interface we constrain outputs to a single option letter. \citep{talmor2019commonsenseqa}
\begin{quote}
\footnotesize\ttfamily
Q: \{question\}\newline
(A) \{choiceA\}\newline
(B) \{choiceB\}\newline
(C) \{choiceC\}\newline
(D) \{choiceD\}\newline
(E) \{choiceE\}\newline
Answer (A/B/C/D/E):
\end{quote}

\paragraph{HaluEval / HalluLens / FActScore template.}
We keep a minimal context+prompt+answer format aligned with common benchmark usage. \citep{li2023halueval,bang2025hallulens, min2023factscore}
\paragraph{HaluEval-QA template (evidence-grounded QA).}
\begin{quote}
\footnotesize\ttfamily
Knowledge (may be empty):\newline
\{knowledge\}\newline\newline
Question:\newline
\{question\}\newline\newline
Answer:
\end{quote}

\paragraph{HaluEval-Dialogue template (multi-turn).}
\begin{quote}
\footnotesize\ttfamily
Conversation:\newline
\{dialogue\}\newline
Assistant:
\end{quote}

\paragraph{HalluLens PreciseWikiQA / LongWiki template (wiki-grounded QA).}
\begin{quote}
\footnotesize\ttfamily
Context:\newline
\{wiki\_context\}\newline\newline
Question:\newline
\{question\}\newline\newline
Answer (do not invent facts not supported by the context):
\end{quote}

\paragraph{HalluLens NonExistentRefusal template (refuse if non-existent).}
\begin{quote}
\footnotesize\ttfamily
User request:\newline
\{prompt\}\newline\newline
Answer. If the entity or requested information does not exist or cannot be verified, say you do not know rather than guessing:
\end{quote}

\paragraph{FActScore biography prompt (official query form).}
\begin{quote}
\footnotesize\ttfamily
Tell me a bio of \{entity\}.
\end{quote}

\section{Input Degradation Protocol}
\label{sec:app-degradation}

We perturb the \emph{user prompt only} and keep labels/answers unchanged.
We follow prior robustness-style token deletion, and implement token replacement (mixup) as random word replacement from a fixed vocabulary. \citep{ishibashi2023evaluating, xie2017data}
\begin{itemize}[leftmargin=1.25em]
  \item \textbf{Token deletion}: delete $r\in\{0.05,0.10,0.15\}$ of word tokens uniformly at random.
  \item \textbf{Token mixup}: replace $r\in\{0.05,0.10,0.15\}$ of word tokens with content words sampled from a fixed vocabulary.
\end{itemize}
Perturbations are deterministic under a fixed seed and applied consistently across methods.

\section{Efficiency Measurement Protocol}
\label{sec:app-efficiency}

We report:
\begin{itemize}[leftmargin=1.25em]
  \item \textbf{Opt.\ time and internal calls (APO)}: offline wall-clock time and the number of optimizer/critic/evaluator calls required to produce one optimized prompt per benchmark.
  \item \textbf{Specialist calls (POaaS)}: average number of specialist calls per query (often $<1$ due to skipping).
  \item \textbf{Added prompt tokens}: additional \emph{input} tokens prepended at inference time (tokenized with the target tokenizer), distinct from the fixed 512-token \emph{generation} cap.
\end{itemize}

\end{document}